\documentclass{edm_article}
\usepackage{xcolor}
\usepackage{amsmath}
\usepackage{graphicx}
\usepackage{booktabs}
\usepackage{subcaption}

\newcommand{\vtirtdirloc}{VTIRT\textsubscript{dir-loc}}
\newcommand{\vtirtdirseq}{VTIRT\textsubscript{dir-s2s}}
\newcommand{\vtirttrans}{VTIRT\textsubscript{transfer}}

\newtheorem{theorem}{Theorem}

\begin{document}

\title{Variational Temporal IRT: Fast, Accurate, and Explainable Inference of Dynamic Learner Proficiency}

\numberofauthors{4}
\author{
    \alignauthor Yunsung Kim\\
    \affaddr{Stanford University}\\
    \email{yunsung@stanford.edu}
    \alignauthor Sreechan Sankaranarayanan\\
    \affaddr{Amazon.com LLC}\\
    \email{sreeis@amazon.com}
    \and
    \alignauthor Chris Piech\\
    \affaddr{Stanford University}\\
    \email{piech@cs.stanford.edu}
    \alignauthor Candace Thille\\
    \affaddr{Stanford University}\\
    \email{cthille@stanford.edu}
}
\maketitle

\begin{abstract}
    Dynamic Item Response Models extend the standard Item Response Theory (IRT) to capture temporal dynamics in learner ability. While these models have the potential to allow instructional systems to actively monitor the evolution of learner proficiency in real time, existing dynamic item response models rely on expensive inference algorithms that scale poorly to massive datasets. In this work, we propose Variational Temporal IRT (VTIRT) for fast and accurate inference of dynamic learner proficiency. VTIRT offers orders of magnitude speedup in inference runtime while still providing accurate inference. Moreover, the proposed algorithm is intrinsically interpretable by virtue of its modular design. When applied to 9 real student datasets, VTIRT consistently yields improvements in predicting future learner performance over other learner proficiency models. 
\end{abstract}




\section{Introduction}

Evaluating the proficiency of a student is a fundamental task in education, and decades-long research in psychometrics have developed accurate probabilistic models to measure evidence of proficiency from student behaviors~\cite{sawyer2005cambridge}. Item Response Theory (IRT) is the most well-known and widely applied probabilistic approach to proficiency modeling, which recognizes each response as a joint outcome of item features and student proficiency~\cite{van1997handbook}, and allows a single proficiency value per student to be estimated from responses to multiple assessment items. 

However, in many routine aspects of educational practice, instructors and computer-based learning systems often use assessments more actively to assist learning rather than to evaluate learner proficiency post-hoc. Such assessments are referred to as \emph{formative assessments} and are used not only to track student learning and make appropriate instructional interventions, but also to allow learners to practice their knowledge and skills, and make necessary self-corrections \cite{sawyer2005cambridge}. When learning occurs alongside assessment, learner proficiency is longitudinal rather than inert, and the assumption of static proficiency makes standard IRT less suitable as a model of proficiency measurement.

Dynamic Item Response models~\cite{martin2002dynamic,imai2016fast} mitigate this issue by removing the assumptions of static ability and instead allowing it to stochastically change over time, but existing inference methods rely on expensive iterative algorithms with heavy runtime bottleneck. These methods scale poorly to massive datasets, which can be critical since in most use cases of dynamic proficiency modeling (e.g., learner proficiency monitoring), evaluation often needs to take place \emph{real-time} to monitor the evolution of learner proficiency. This means that the expensive cost of inference must be incurred not just once, but multiple times over the course of a learner's learning experience.

In this paper, we develop Variational Temporal IRT (VTIRT), a fast and accurate framework for inferring dynamic learner proficiency over time. VTIRT is based on the idea of amortized variational inference~\cite{kingma2013auto}, a fast approximate Bayesian inference framework for complex probabilistic models. The resulting algorithm infers the ability trajectory of a learner by first making \emph{local} ability estimates in the form of a Gaussian distribution based on the item and response at each timestep (which we call the ``\emph{ability potentials}''), then aggregating these ability estimates across time in an intuitive fashion. In particular, our work delivers the following key innovations\footnote{Our public implementation of VTIRT based on PyTorch and Pyro~\cite{bingham2019pyro} is available online in the following repository: \texttt{https://github.com/yunsungkim0908/vtirt}}:

\begin{itemize}
    \item \textbf{Interpretable Inference for Dynamic IRT.} VTIRT allows the use of a structured probabilistic inference algorithm for sequence models through the notion of \emph{ability potentials}, a form of conjugate potentials described in~\cite{johnson2016composing}. We concretely derive VTIRT in detail and discuss the explainability of each of its components.

    \item \textbf{Fast and Accurate Inference.} Our proposed inference algorithm yields orders of magnitude speedup in inference runtime compared to existing inference algorithms while maintaining accurate inference.


    \item \textbf{Applications to Real World Datasets.} We apply our inference algorithm to 9 real student datasets. VTIRT consistently yields improvements in predicting future learner performance compared to other existing proficiency models.
\end{itemize}

\section{Related Works}
\label{sec:related}


Many studies~\cite{van2008categorical,studer2012incorporating,ekanadham2017t,wang2013bayesian,weng2018real,martin2002dynamic} have investigated dynamic extensions of IRT that allow learner proficiency to vary over time. A common structure shared by these approaches is that student ability is assumed to follow a random walk:
\begin{equation*}
    \theta_{\ell,t} = \theta_{\ell,t-1} + \varepsilon_{\ell,t},
\end{equation*}
where $\varepsilon_{\ell,t}$ models a stochastic change in ability (often a zero-mean Gaussian). 
\cite{ekanadham2017t} finds a coarse approximation to the posterior distribution of per-time-step ability by ignoring the cross-temporal dependencies in the likelihood function while assuming knowledge of the item parameters.
\cite{martin2002dynamic} and~\cite{wang2013bayesian} use Markov Chain Monte Carlo (MCMC) methods~\cite{brooks2011handbook} to estimate the unknown ability and item parameters. These methods draw samples asymptotically from the true posterior distribution conditioned on the observed responses, but the convergence of MCMC can be slow.
On the other hand, \cite{imai2016fast} and~\cite{weng2018real} use Expectation-Minimization (EM) to iteratively estimate the dynamic item response parameters. In particular, \cite{imai2016fast} uses variational EM (VEM) to estimate the parameters of a distribution that closely approximates the true posterior distribution conditioned on the observed response. Although generally faster than MCMC-based methods, VEM methods still require costly iterative updates. 

Closely related to the task of dynamic proficiency modeling is \emph{knowledge tracing}~\cite{corbett1994knowledge,piech2015deep}, which attempts to trace the knowledge of learners over time and accurately predict future performance. While Markov chain-based methods such as BKT~\cite{corbett1994knowledge} allow proficiency to be numerically measured through the estimated probability of being at a ``proficient'' state, the knowledge state representations of neural network-based knowledge tracing models~\cite{piech2015deep} are not readily comparable or interpretable. Logistic regression knowledge tracing models offer simple and interpretable alternatives to neural network-based models. BestLR~\cite{gervet2020deep} and LKT~\cite{pavlik2021logistic} belong to this family of methods and use the number of correct and incorrect attempts as input features, while DAS3H~\cite{choffin2019das3h} additionally embeds explicit representations of learning and forgetting over spans of time. VTIRT produces numerical representations of learner proficiency that are comparable by design across learners and across time, and its interpretable inference is also sensitive to the features of the attempted items.

Amortized variational inference has been used in~\cite{wu2020variational} to develop VIBO for standard IRT. VIBO and its relationship to VTIRT are further discussed in Section~\ref{sec:vibo}.

\section{Variational Inference Review}
\label{sec:vi}

Variational inference is a Bayesian framework for efficiently inferring unobserved variables in complex probabilistic models. In this setting, observations are modeled as samples from some underlying probability distribution (called the \emph{generative model}) where some of the random variables (denoted $r$) are observed, and the remaining \emph{latent} variables (denoted $z$) are unobserved. 
The goal of Bayesian inference then is to infer the latent random variables by finding the posterior distribution $p(z|r)$ given our knowledge of the likelihood distribution $p(r|z)$ and the prior distribution $p(z)$. This has the effect of ``updating'' the prior belief $p(z)$ with the observations to obtain the posterior belief $p(z|r)$. 


For complex generative models, the posterior distribution $p(z|r)$ is often intractable to compute exactly. Variational inference is one way of doing \emph{approximate} posterior inference that treats inference as an optimization problem, where we find the distribution $q(z)$ that is closest to the true posterior $p(z|r)$ from a more constrained (yet rich) family of distributions $\mathcal{Q}$ of our choice. This is achieved by maximizing an objective called ``Evidence Lower BOund'' (ELBO) for the observation $r$ with respect to $q$
\begin{equation}
    \mathcal{L}(q) \triangleq \mathbb{E}_{q(z)}\left[{\frac{\log p(r|z)p(z)}{\log q(z)}}\right],
    \label{eq:elbo}
\end{equation}
which is equivalent to minimizing the Kullback-Leibler divergence between $q(z)$ and $p(z|r)$\footnote{In fact, if $\mathcal{Q}$ includes the true posterior, then the $q$ that achieves optimality will exactly be the the true posterior.} due to the following equality:
\begin{equation*}
    \mathcal{L}(q) + KL\left({q(z) \| p(z|r)}\right) = \log p(r) \equiv \text{Constant w.r.t }q.
\end{equation*}


\paragraph{Amortized Inference}

What we just described is how VI works for a single observation. If we have a set of multiple i.i.d. observations sampled from the data-generating distribution $p_{\mathcal D}$ (which will be equal to the marginal distribution $p(r)$ if our generative model is correctly chosen), then finding the approximate posterior is equivalent to the following optimization problem
\begin{equation}
    \arg\max_q \mathcal{L}(q) \triangleq \mathbb{E}_{p_{\mathcal D}(r)}\left[{
        \mathbb{E}_{q_r(z)}\left[{\frac{\log p(r,z)}{\log q_r(z)}}\right]
    }\right]
\end{equation}
where we find one variational posterior factor $q_r$ for each observation $r$. As the number of observations grows, however, finding $q_r$ for each observation can quickly become highly inefficient. \emph{Amortized Variational Inference}~\cite{gershman2014amortized} tries to avoid this issue by \emph{learning a mapping} $\phi(r)$ (also called the ``recognition model'') that maps observations to the parameters of the corresponding posterior distribution, rather than inferring each approximate posterior on the fly. By training a good recognition model ahead of time based on data and using it to retrieve the posterior distribution almost instantaneously at inference time, the cost of per-observation inference can be \emph{amortized}~\cite{gershman2014amortized}. Now we can choose the recognition model from a highly expressive family of functions (e.g., a neural network) and optimize the recognition model instead:
\begin{equation}
    \arg\max_{\phi}\mathcal{L}(\phi) \triangleq 
    \arg\max_{\phi}\mathbb{E}_{p_{\mathcal D}(r)}\left[{
        \mathbb{E}_{q_{\phi(r)}(z)}\left[{\frac{\log p(r,z)}{\log q_{\phi(r)}(z)}}\right]
    }\right].
\end{equation}

\section{The VTIRT Framework}
\label{sec:vtirt}

\begin{figure*}[t]
\centering
\begin{subfigure}{.45\textwidth}
  \centering
  \includegraphics[width=0.9\linewidth]{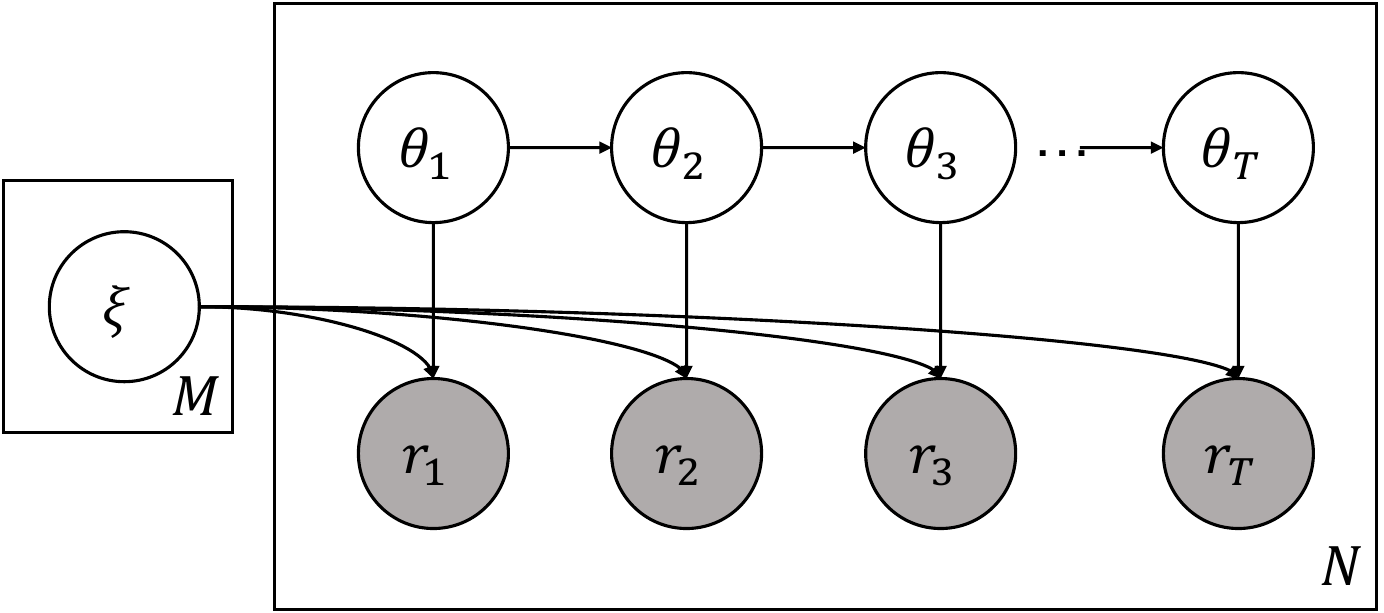}
  \Description{This figure presents VTIRT's generative model.}
  \caption{VTIRT's Generative Model}
  \label{fig:graph1}
\end{subfigure}%
\quad
\begin{subfigure}{.45\textwidth}
  \centering
  \includegraphics[width=0.9\linewidth]{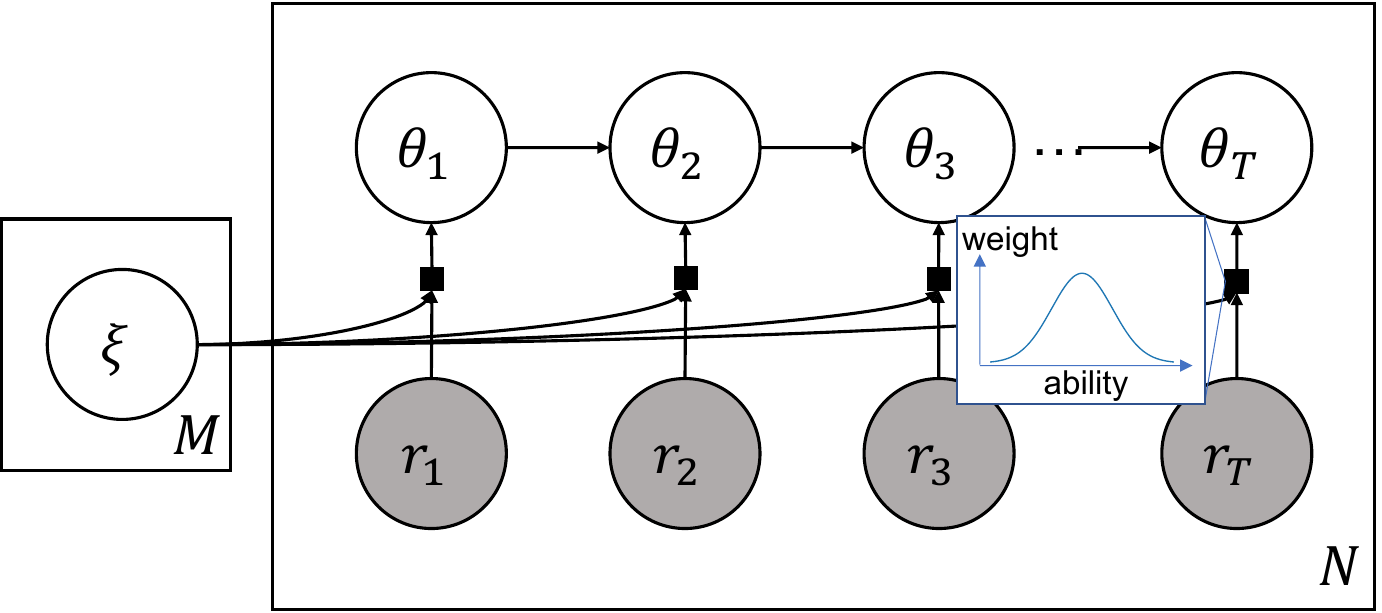}
  \Description{This figure presents VTIRT's inference model, or the variational posterior family.}
  \caption{VTIRT's Inference Model}
  \label{fig:graph2}
\end{subfigure}
\caption{Graphical model view of VTIRT's generative model and inference model. Shaded nodes indicate observed variables, and arrows denote the direction of dependence. Squares denote the ability potentials in the form of a Gaussian density.}
\Description{VTIRT's generative model has dependency arrows going from item parameters and the ability variables into the responses at each timestep. VTIRT's inference model has dependency arrows going from the responses and item parameters into a potential function (drawn as a shaded square) that has another arrow going into the ability variable at each timestep. Both the generative and inference models have arrows that go from the ability variable at one timestep to the ability variable at the next timestep.}
\label{fig:graph}
\end{figure*}

Based on the ideas of variational inference introduced earlier, we are now ready to describe the generative model and the inference algorithm that together comprise the VTIRT framework. The main intuition behind VTIRT's generative model is to incorporate temporality into IRT in a way similar to~\cite{ekanadham2017t,wilson2016back}. Our framework, however, offers the additional flexibility to use \emph{any} form of the item characteristic function - potentially with learnable parameters - whereas prior methods are constrained to a specific functional form.

\subsection{The Temporal Ability Model}
\label{sec:gen_model}

In our generative model (Figure~\ref{fig:graph1}), we assume that the response $r_{\ell,t}$ of learner $\ell$ at timestep $t$ is determined by 2-parameter IRT,
\begin{equation}
    p\left({r_{\ell,t}|\theta,a,d}\right) = f\left({a_{q_{\ell,t}}\left({\theta_{\ell,t} - d_{q_{\ell,t}}}\right)}\right),
    \label{eq:likelihood}
\end{equation}

where $q_{\ell,t}$ denotes the assessment item, $\theta_{\ell,t}\in [-\infty,\infty]$ denotes the ability of learner $\ell$ at timestep $t$, $a_q$ and $d_q$ each denote the discrimination and difficulty of assessment item $q$, and $f$ denotes the linking function. To infuse temporality, we take an approach similar to~\cite{ekanadham2017t,wilson2016back} and impose an additional assumption that a learner's ability is sampled from a random walk with Gaussian noise, also called a Wiener Process:
\[
    \theta_{\ell,t+1}|\theta_{\ell,t} \sim \mathcal{N}(\theta_{\ell,t}, \sigma_\theta^2), \quad \theta_{\ell,0} \sim \mathcal{N}(0, \sigma_\theta^2).
\]
This is an instance of a more general Linear Gaussian model (LGM)
\begin{equation}
    \theta_{\ell,t+1}|\theta_{\ell,t} 
    \sim
    \mathcal N(\alpha_{\ell,t}\cdot\theta_{\ell,t}+\beta_{\ell,t}, s_{\ell,t})
    \label{eq:lgm}
\end{equation}
where the scale, bias, and standard deviation parameters are set to $(\alpha_{\ell,t},\beta_{\ell,t},s_{\ell,t})=(1,0,\sigma_\theta).$\footnote{To allow for a fully Bayesian treatment, we also impose a Gaussian prior distribution on the item parameters: $a_q \sim \mathcal{N}(1,\sigma_a^2)$, and $d_q \sim \mathcal{N}(0, \sigma_d^2)$.}

The most popular choice for the linking function is the sigmoid function for 2 parameter logistic (2PL) IRT and Gaussian CDF for 2 parameter O-give (2PO) IRT. We will use 2PL as our modeling choice in our experiments considering its popularity~\cite{van1997handbook}. It is important to note, however, that VTIRT makes \emph{no assumption} about the linking function $f$ as long as $f$ is differentiable. Moreover, we can straightforwardly extend the model to admit a parameterized custom linking function $f_\psi$ which we can learn from data. A similar approach in~\cite{wu2020variational} has proven to yield better fit and higher predictive performance in the case of standard IRT, and we leave this extension to future research. This is in contrast to prior algorithms~\cite{ekanadham2017t,wilson2016back} that become intractable for any linking functions other than a Gaussian CDF.

\subsection{Choosing the Variational Family $\mathcal{Q}$}

To do inference on our generative model, we first need to choose the variational family $\mathcal Q$. We will choose $\mathcal Q$ to be the family of distributions that factorize as follows:
\begin{equation}
    q(\xi,\theta;r) = q(\xi)q(\theta|\xi,r) = \left({\prod_q q(\xi_q)}\right)\left({\prod_\ell q(\theta_\ell|\xi,r)}\right),
\end{equation}
where we have used the shorthand notation  $\xi_q = (a_q, d_q)$ to denote the features of the assessment item $q$. Since we are interested in inferring the temporal trajectory of abilities, we will choose $q(\theta|\xi,r)$ to be a Linear Gaussian Model just as its prior $p(\theta)$, and also choose $q(\xi)$ to be Gaussian. More precisely, we define $q(\theta|\xi,r)$ such that
\begin{equation}
    \theta_{\ell,t+1}|\theta_{\ell,t},\xi,r_\ell \sim \mathcal N\left({\alpha_{\ell,t}\cdot\theta_{\ell,t} + \beta_{\ell,t}, s_{\ell,t}}\right)
\end{equation}
whose scale $\alpha_{\ell,t}$, bias $\beta_{\ell,t}$, and standard deviation $s_{\ell,t}$ parameters are dependent on $\xi$ and $r_\ell$. Recalling the variational lower bound from Equation~\eqref{eq:elbo}, our objective becomes
\begin{equation}
    \mathcal{L}(q) = \mathbb{E}_{q(\xi)q(\theta|\xi,r)}\left[{
        \frac{p(\xi)p(\theta)p(r|\xi,\theta)}
        {q(\xi)q(\theta|\xi,r)}
    }\right].
    \label{eq:vtirt_elbo}
\end{equation}

Since the parameters $\alpha_\ell$, $\beta_\ell$ and $s_\ell$ are dependent on the item parameters $\xi$ and observed responses $r_\ell$, it is tempting to apply the idea of amortized inference from Section~\ref{sec:vi} directly and model these parameters using learnable mappings. One such approach that we call \textbf{\vtirtdirloc} is to map the transition parameters at each timestep $1 \leq t \leq T$ based on the item parameters and responses from that timestep
\begin{equation}
    \alpha_{\ell,t}, \beta_{\ell,t}, s_{\ell,t}
    = \phi\left({\xi_{q_{\ell,t}}, r_{\ell,t}}\right).
\end{equation}
While this approach is modular and its recognition model is low-dimensional and visualizable, its parameter estimates are not allowed to depend on responses \emph{through time}, which may produce sub-optimal fit as we will later demonstrate through experiments. To allow dependence through time, we could instead choose to use a sequence-to-sequence recognition network (such as an LSTM network) to estimate the parameters for all time-steps at once using the entire sequence of responses:
\begin{equation}
    \alpha_{\ell,1:T}, \beta_{\ell,1:T}, s_{\ell,1:T}
    = \phi\left({\xi_{q_{\ell,1:T}}, r_{\ell,1:T}}\right).
\end{equation}
We call this approach \textbf{\vtirtdirseq}. While this uses a more expressive mapping, the increased complexity comes at the cost of interpretability and potentially a greater demand for more training data and long input sequences.

To mitigate this trade-off, we instead opt for an approach that is both modular enough to yield interpretability and yet also allows parameter estimates to depend on the responses through time.

\subsection{VTIRT's Inference Algorithm}
\label{sec:vtirt_potentials}

To describe our main inference method \textbf{VTIRT}, we first draw our attention to the following property about Linear Gaussian Models and Wiener processes, which will be foundational to our proposed method (See Appendix~\ref{sec:proof} for the proof):
\begin{theorem}
    Let $p(\theta_{1:T})$ be a Wiener process with standard deviation $\sigma_\theta$ and $q(\theta_{1:T})$ be a probability distribution defined as 
    \begin{equation}
        q(\theta_{1:T}) \propto p(\theta_{1:T})\prod_{t=1}^T \exp\left\{{\left({
        \frac{\theta_{t} - \mu_t}
        {\sigma_t}
        }\right)^2}\right\},
        \label{eq:gaussian_potential}
    \end{equation}
    for real numbers $\mu_{1,...,T}$ and $\sigma_{1,...,T}$.

    Then, $q(\theta_{1:T})$ is a Linear Gaussian Model\footnote{For notational convenience, we will use $\theta_0=0$}
    \begin{equation}
        \theta_t | \theta_{t-1} \sim \mathcal{N}(
            \widetilde \mu_t, \widetilde \sigma_t
        )
        \label{eq:q_conditional}
     \end{equation}
     with
     \begin{equation}
         \widetilde \mu_t = \left({\frac{
            \lambda_\theta\theta_{t-1}
            + \lambda_t\mu_t 
            + (\rho_{t+1}\lambda_\theta)\tau_{t+1}
         }{
            \lambda_\theta + \lambda_t 
            + (\rho_{t+1}\lambda_\theta)
         }}\right)
    \label{eq:mu_tilde}
    \end{equation}
    and
    \begin{equation}
        \widetilde \sigma_t = \sigma_\theta\sqrt{1-\rho_{t+1}},
    \label{eq:sigma_tilde}
    \end{equation}
    where $\lambda_\theta=1/\sigma^2_\theta$ and $\lambda_t=1/\sigma^2_t$ denote precisions and parameters $\rho_t$ and $\tau_t$ are defined recursively as
    \begin{equation}
        \rho_t = \left({\frac{
            \lambda_t + (\rho_{t+1}\lambda_\theta)
        }{
            \lambda_\theta
            + \lambda_t + (\rho_{t+1}\lambda_\theta)
        }}\right), \rho_{T+1} = 0
    \label{eq:rho_t}
    \end{equation}
    and
    \begin{equation}
        \tau_t = \left({\frac{
            \lambda_t\mu_t + (\rho_{t+1}\lambda_\theta)\tau_{t+1}
        }{
            \lambda_t + (\rho_{t+1}\lambda_\theta)
        }}\right), \tau_{T+1}=0.
    \label{eq:tau_t}
    \end{equation}
\label{THM:GAUSSIAN_POTENTIAL}
\end{theorem}

In Equation~\eqref{eq:gaussian_potential}, we are defining $q$ by attaching \emph{local ``ability potentials''} to the prior distribution $p$, where each potential term is in the form of a Gaussian density with mean $\mu_t$ and variance $\sigma^2_t$. These potentials could be understood as local ``beliefs'' about the ability in the form of Gaussian distributions, judged solely based on the item features and learner response at the current timestep.

\begin{figure}[t]
    \centering
    \includegraphics[width=0.55\linewidth]{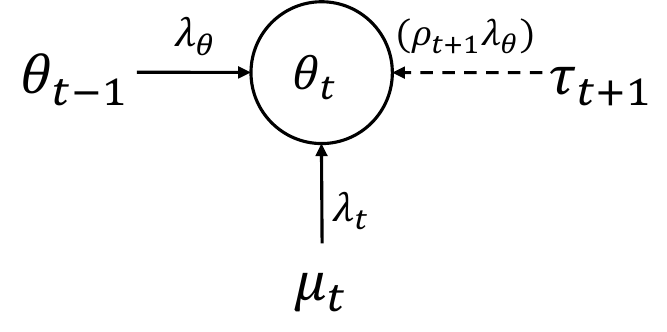}
    \caption{Schematic of VTIRT's inference at each timestep.}
    \Description{Information is aggregated from different points in time to make inference for ability at a each time step.}
\label{fig:inference}
\end{figure}

These potentials are combined across time with the prior distribution $p(\theta)$. The resulting $\theta_t$ follows a Gaussian distribution whose mean is a \emph{weighted average} of the following 3 values that each represent information from different points in time (Figure~\ref{fig:inference}): (1) $\theta_{t-1}$ of the previous timestep, (2) the local potential mean $\mu_t$ of the current timestep, and (3) the ``future potential aggregate'' $\tau_{t+1}$ that recursively aggregates potentials backwards from future timesteps via weighted averaging (Equation~\eqref{eq:tau_t}). Each value is weighted proportionally to the \emph{precision} (or ``inverse uncertainty'') associated with it\footnote{$\rho_t\lambda_\theta$ can be viewed as the \emph{effective precision} of the information coming from future timesteps.}, so the term with the lowest uncertainty contributes most to the resulting mean.

Therefore, Theorem~\ref{THM:GAUSSIAN_POTENTIAL} suggests a way to aggregate local ability estimates (Gaussian ability potentials) across timesteps using the global prior structure of the generative model. This motivates us to choose the following family of distributions for our variational factor $q(\theta)$ (Figure~\ref{fig:graph2}):
\begin{equation}
    q(\theta_\ell) \propto p(\theta_\ell)\prod_{t}\exp\left\{{\left({
        \frac{\theta_{\ell,t} - \mu(\xi_{q_{\ell,t}}, r_{\ell,t})}
        {\sigma(\xi_{q_{\ell,t}}, r_{\ell,t})}
        }\right)^2}\right\},
\end{equation}
where $\mu(\cdot,\cdot)$ and $\sigma(\cdot,\cdot)$ are parameterized functions (e.g., feed-forward neural networks) that play the role of the recognition model. We refer to the resulting inference algorithm as \textbf{VTIRT}.

\subsection{Conjugate Potentials and Variational IRT}
\label{sec:vibo}

VTIRT can be considered as a special case of using conjugate potential functions~\cite{johnson2016composing} to conduct approximate Bayesian inference, which allows intuitive and efficient inference algorithms designed for conditionally conjugate models to be used even when the model violates conjugacy. Specifically, the ability potentials in VTIRT enable efficient computation of variational posterior factors using a fast forward-backward inference algorithm for Linear Gaussian Models outlined in Theorem~\ref{THM:GAUSSIAN_POTENTIAL}.

VIBO~\cite{wu2020variational}, an amortized variational inference algorithm for standard IRT, also belongs to this family of methods. In VIBO, the variational posterior distribution for ability is a Product-of-Experts where each ``expert'' component is a Gaussian distribution that depends locally on the response and item parameters from each timestep. These ``experts'' are also a form of conjugate potentials that allow variational posterior factors to be computed in closed-form. 

This leads to several commonalities in both frameworks. Both use the same set of learnable parameters - the Gaussian posterior parameters ($\mu_{a_q}$, $\mu_{d_q}$, $\sigma^2_{a_q}$, $\sigma^2_{d_q}$) for each item $q$, and two recognition function components $\mu(\cdot,\cdot)$ and $\sigma(\cdot,\cdot)$ - and make inference by aggregating local ability potentials. While VIBO aggregates the conjugate potentials into a single univariate distribution over ability through a Product-of-Experts, VTIRT aggregates them into a Linear Gaussian Model based on Theorem~\ref{THM:GAUSSIAN_POTENTIAL}. In Section~\ref{sec:eval}, we will demonstrate through experiments that this difference in aggregation leads to VTIRT's performance improvement.

\section{Evaluation}
\label{sec:eval}

We will now demonstrate that VTIRT achieves orders of magnitude faster inference than existing methods without compromising inference quality while also providing an interpretable structure. Experiments with real student data will also demonstrate that VTIRT yields a better fit to student behaviors than other learner proficiency models. We first describe the 2 datasets we used for our experiments.

\subsection{Datasets}

\begin{table}[t]
    \centering
    \caption{Statistics of the Workspace Learning Dataset}
\begin{tabular}{cccc}
\toprule
Course Name &  Items &  Learners &  Interactions \\
\midrule
Interviewing 1       &     89 &   79,808 &    5,458,576\\
Interviewing 2       &     12 &   10,536 &      120,388\\
Design Thinking      &     12 &   45,369 &      458,232\\
Software Development &      8 &   10,277 &       80,137\\
Document Writing     &     13 &   20,043 &      233,175\\
Management A-1       &     28 &   10,154 &      247,674\\
Management A-2       &     16 &   14,673 &      198,720\\
Management B-1       &     14 &   21,293 &      281,844\\
Management B-2       &     14 &   15,254 &      206,108\\
\bottomrule
\end{tabular}

    \label{tab:data_stats}
\end{table}

\subsubsection{Synthetic Dataset} Using a simulated dataset enables us to test our algorithm under various hypothetical circumstances. We use VTIRT's generative model to simulate a set of learners responding to assessment items in an arbitrary order. For each learner, we first choose a random permutation of assessment items to simulate learners responding to assessment items in arbitrary order. Responses to these items are sampled based on the generative model defined in Section~\ref{sec:gen_model}. This gives us access to the ground-truth item features and ability values that are otherwise unobtainable in real-world datasets. We set $\sigma_\theta=0.25$ and $\sigma_a = \sigma_d = 1$ and vary the number of learners and the number of items.

\subsubsection{Real Student Dataset: Workplace Learning} This dataset contains anonymized learner responses to a series of assessment questions in workplace learning courses taken by employees of a company. Each interaction record consists of (1) the ID of the assessment item (question), (2) ID of the learner, (3) correctness of the attempt, and (4) the knowledge components\footnote{Most courses had 2-4 knowledge components.} with which each assessment item is associated (of which there could be multiple). Learners with fewer than 5 interactions throughout the course were omitted, and if there were multiple attempts to a question, only the first attempts were retained. A set of summary statistics for this dataset is presented in Table~\ref{tab:data_stats}.

\subsection{Fast and Accurate Inference}
\label{sec:synth_eval}

\begin{figure*}[t]
    \centering
    \includegraphics[width=0.9\linewidth]{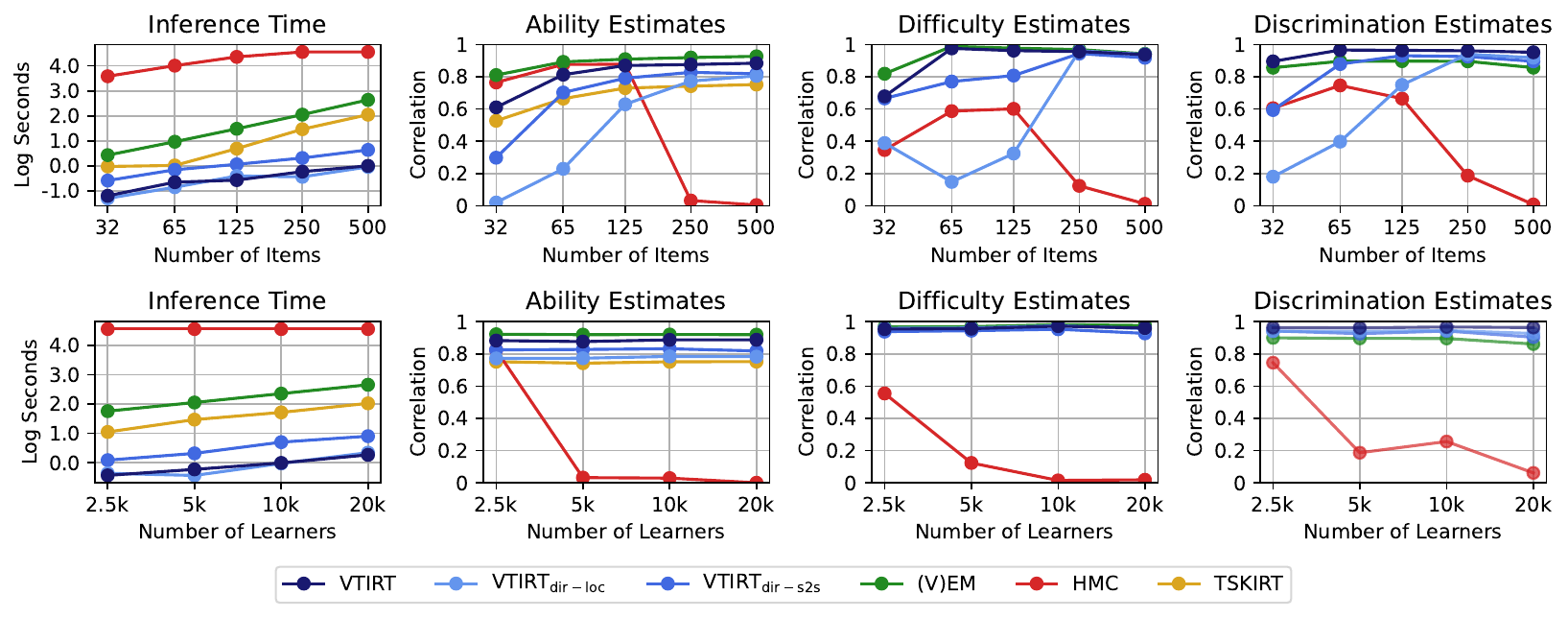}
    \caption{Performance on the synthetic dataset. Inference time was capped at 10 hours.}
    \Description{The main text has a detailed analysis and description of this performance plot.}
\label{fig:synth}
\end{figure*}

The most important quality of an inference algorithm is its capacity to promptly and reliably recover the unobserved variables based on past observations. The synthetic dataset allows us to measure this by comparing the computational runtime of a single instance of inference and computing the correlation of the inferred ability and item features against the known ground-truth values.

We implemented the 3 variants of VTIRT (\vtirtdirloc, \vtirtdirseq, and VTIRT) along with 3 existing baseline inference methods - Variational EM (VEM), MCMC using Hamiltonian Monte Carlo\footnote{Hamiltonian Monte Carlo~\cite{betancourt2017conceptual,hoffman2014no} is an efficient MCMC algorithm for continuous state spaces.} (HMC), and TSKIRT~\cite{ekanadham2017t} - and used these algorithm to recover the latent ability values and item features for all learners and trials, based on the responses from all timesteps. (See Appendix~\ref{sec:exp_details} for more details about the methods and the experiment.) We varied the number of items from 32 to 500 while fixing the number of learners to 5000, then varied the number of learners from 2,500 to 20,000 while fixing the number of items to 250.

Figure~\ref{fig:synth} plots the inference time and Pearson correlations of the model estimates with the ground-truth values. Most notably, all 3 variants of VTIRT are \emph{orders of magnitude} faster than other inference methods. Moreover, VTIRT consistently yields the best discrimination estimates. Except when there are few items, the difference in the quality of ability and difficulty estimates are also minor compared to VEM (up to 0.07 difference in ability correlation and 0.03 difference in difficulty correlation).

Among all variants of VTIRT, VTIRT using ability potentials consistently outperforms direct amortization. As noted earlier, \text{\vtirtdirloc} ignores temporal dependency in estimating the transition dynamics, while the complexity of \text{\vtirtdirseq} could come at the cost of the need for more training data and long input sequences.

\subsection{Application to Real Student Data}
\label{sec:real_world_eval}

\begin{table*}[t]
    \centering
    \caption{Next-Step Performance Prediction ROC.}
\begin{tabular}{c|ccc|cccc}
\toprule
{} &   IRT &   BKT &  VIBO &  VTIRT &  VTIRT\textsubscript{dir-loc} &  VTIRT\textsubscript{dir-s2s} &  VTIRT\textsubscript{transfer} \\
\midrule
Interviewing 1       & 0.702 & 0.622 & 0.752 &  \textbf{0.762} &                     0.758 &                         0.749 &                       0.756 \\
Interviewing 2       & 0.586 & 0.632 & 0.765 &  \textbf{0.779} &                     0.774 &                         0.772 &                       0.760 \\
Software Development & 0.565 & 0.648 & 0.701 &  \textbf{0.711} &                     0.695 &                         0.667 &                       0.702 \\
Design Thinking      & 0.602 & 0.605 & 0.674 &  \textbf{0.681} &                     0.677 &                         0.646 &                       0.633 \\
Document Writing     & 0.503 & 0.683 & 0.754 &  \textbf{0.770} &                     0.766 &                         0.750 &                       0.746 \\
Management A-1       & 0.518 & 0.639 & 0.717 &  \textbf{0.738} &                     0.734 &                         0.729 &                       0.723 \\
Management A-2       & 0.705 & 0.682 & 0.771 &  \textbf{0.774} &                     0.770 &                         0.766 &                       0.770 \\
Management B-1       & 0.570 & 0.582 & 0.734 &  \textbf{0.741} &                     0.739 &                         0.730 &                       0.735 \\
Management B-2       & 0.733 & 0.602 & 0.766 &  \textbf{0.770} &                     0.766 &                         0.765 &                       0.766 \\
\bottomrule
\end{tabular}

    \label{tab:pred_result}
\end{table*}

We now compare VTIRT with other proficiency models in modeling real student data. Since we do not have access to the ground-truth learner ability in reality, our evaluation on real student data must be based on a related proxy metric. As a proxy, we will focus on the task of predicting the \emph{next step} response correctness of learners based on the model's \emph{current} ability estimates and item features.\footnote{Since the items in each course were associated with different knowledge components, we estimated learner ability for each knowledge component separately. Prediction on each item was made based on the ability averaged across the knowledge components associated with that item.}

We compared the predictive performance of VTIRT against the following baseline: IRT, BKT, VIBO\cite{wu2020variational}\footnote{To adopt VIBO to a sequential estimation setting, we computed the ability estimates at each timestep separately using the responses prior to that timestep.}, \vtirtdirloc, and \vtirtdirseq.\footnote{We used the popular MIRT package in R for the IRT baseline, and the implementation from the pyBKT package~\cite{badrinath2021pybkt} for the BKT baseline. Since VTIRT and VIBO's estimates take the form of a probability distribution, we used the mean of the distribution as the model's point-estimate and fed it as input to the 2PL IRT likelihood function in Equation~\eqref{eq:likelihood} to compute the predicted probability of correctness.} To study the effect of VTIRT's forward-backward inference algorithm, we also analyzed the performance of a variant of VTIRT we call \textbf{\vtirttrans} in which we train the recognition networks using VIBO and perform inference using VTIRT's inference algorithm.

Table~\ref{tab:pred_result} reports the average AUROC on this prediction task over a 5-fold cross-validation, where the learners were split into 5 equally-sized splits. These results suggest the following observations:
\begin{description}
    \item[VTIRT consistently outperforms other proficiency models.] \hfill\\
    VTIRT achieves up to 2.1 AUROC point advantage in comparison to the best performing baseline, VIBO. As VIBO and VTIRT share the same parameterization scheme, the increased performance is attributable to the VTIRT framework.

    \item[Ability potentials are more effective than direct amortization.] 
    VTIRT using ability potentials outperforms both the local and sequence-to-sequence direct amortization variants. It is interesting to note that local direct amortization also outperformed LSTM-based sequence-to-sequence direct amortization in all courses, which may be due to relatively short sequence length per knowledge component.

    \item[VTIRT's training mechanism is critical to its performance.] \hfill\\
    Since VTIRT and VIBO have the same parameterization schemes, it is natural to ask whether VTIRT's sequential training could be replaced with VIBO's parallelizable training without much loss in performance. Comparing the performance of \vtirttrans with VTIRT, we see that VTIRT's training mechanism is crucial to the enhanced performance, and \vtirttrans often performs far worse than VIBO itself.
\end{description}

\subsection{Interpretability of VTIRT}
\label{sec:interpretability}

VTIRT is a modular algorithm, and by virtue of its structure, all parts of its operations are intrinsically interpretable. The ability estimates are computed from the local ability potentials, following the logic outlined in Section~\ref{sec:vtirt_potentials}. These ability potentials provide ``local beliefs'' of the learner's ability at each timestep in the form of a Gaussian distribution and are aggregated through the forward-backward inference algorithm based on Theorem~\ref{THM:GAUSSIAN_POTENTIAL}.

\begin{figure}[t]
    \centering
    \begin{subfigure}[b]{\linewidth}
        \centering
        \includegraphics[width=0.49 \linewidth]{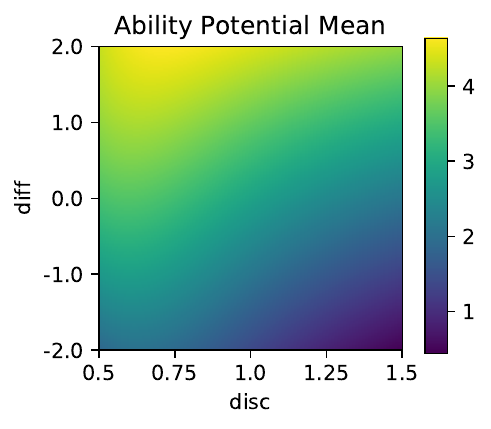}
        \Description{A 2D heatmap of the potential mean function when the response is ``correct.'' See Section~\ref{sec:interpretability} for more detail.}
        \hfill
        \includegraphics[width=0.49\linewidth]{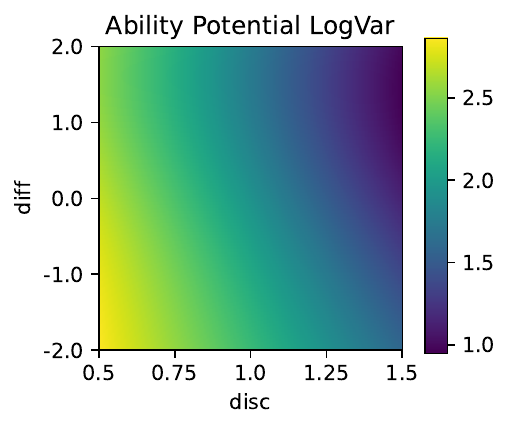}
        \Description{A 2D heatmap of the potential log variance function when the response is ``correct.'' See Section~\ref{sec:interpretability} for more detail.}
        \caption{Correct Response}
    \end{subfigure}
    \vskip\baselineskip
    \begin{subfigure}[b]{\linewidth}
        \centering
        \includegraphics[width=0.49\linewidth]{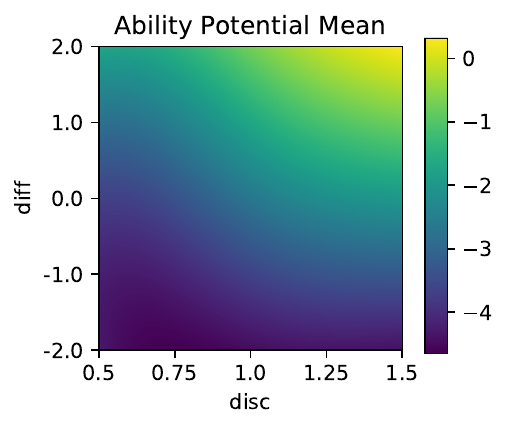}
        \Description{A 2D heatmap of the potential mean function when the response is ``incorrect.'' See Section~\ref{sec:interpretability} for more detail.}
        \hfill
        \includegraphics[width=0.49\linewidth]{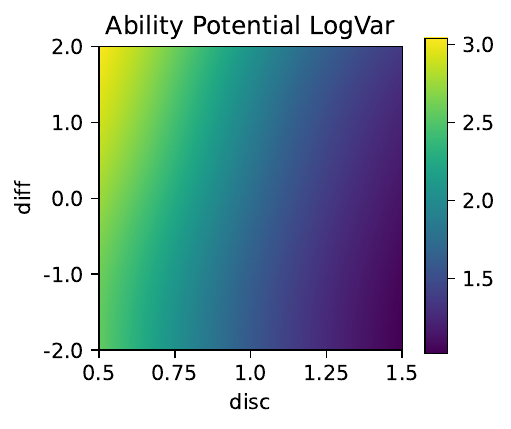}
        \Description{A 2D heatmap of the potential log variance function when the response is ``incorrect.'' See Section~\ref{sec:interpretability} for more detail.}
        \caption{Incorrect Response}
    \end{subfigure}
    \caption{Mean and log variance of the ability potentials as a function of the item correctness and item features.}
    \Description{See Section~\ref{sec:interpretability} for an analysis and description of the visualization of the ability potential function.}
    \label{fig:potentials}
\end{figure}

One of the merits of this potential function is that its dimensions are low enough to be visually analyzed. Figure~\ref{fig:potentials} is a plot of the mean and log variance\footnote{High variance indicates large uncertainty.} of the potential function for the ``Interviewing 2'' course for typical parameter ranges, and its shape aligns with our intuitive expectations of how a learner's response would affect our belief of its ability depending on the item features. In particular,
\begin{itemize}
    \item For assessment items of any difficulty and discrimination, a correct response always yields higher ability estimate than an incorrect response (which can be seen from the range of the color bar).

    \item The uncertainty of the ability estimates are generally lower (so the model is more certain about its estimates) for items with higher discrimination. This aligns with the expectation that high discrimination items are useful for distinguishing learners with different abilities.

    \item Correct responses to high-difficulty items yield potentials with greater mean and lower uncertainty than correct responses to low-difficulty questions (and the opposite for incorrect responses).\footnote{Although it may seem as if correct responses to low-discrimination items yield higher ability estimates because the mean parameter is greater, the overall distribution is in fact flatter and more spread out in general due to higher variance.} 
\end{itemize}

\section{Limitations and Future Work}
\label{sec:discussion}
        
\paragraph{Adaptive and Self-Directed Learning} The key characteristic of VTIRT is its ability to make sequential ability estimates from responses to a set of heterogeneous assessment items. For this reason, we hypothesize that the ideal environment for VTIRT in comparison to other proficiency models is one where learners possess great agency in choosing their learning trajectories, or where the learning trajectories are adjusted adaptively to the performance of the learner. However, most learners in our real student dataset followed similar learning trajectories with little variability, and this hypothesis remains untested. An important direction for future work would be to test our framework in an adaptive or self-directed learning environment.

\paragraph{Modeling Assumptions of VTIRT} One interesting topic for future research is the modeling assumption made by VTIRT. VTIRT's generative model builds on a simple assumption that learner ability starts close to 0 and that the changes in ability are Gaussian with mean 0. Under this generative model, the temporal changes in ability may take on both positive and negative values. While we have shown using real student data that the resulting inference algorithm yields a more accurate fit, research remains to be done to examine how the modeling assumptions could be further improved.

\paragraph{Ability Potential for Atypical Item Parameter Values} In Section~\ref{sec:interpretability}, we visualized in Figure~\ref{fig:potentials} the trained ability potential function for one of the datasets for typical ranges of the item parameter values. Yet, the input to the potential function can be any tuple $\xi = (a,d)$ of unbounded real numbers, and the typical range of input observed during training comprise only a very small subset of this domain. For values of the item parameters outside this typical range, the trained potential function may fail to generalize as a result of sparse training signal and exhibit arbitrary behaviors. Enhancing the generalizability of the potential function and its robustness to extreme values of the item parameters is an exciting direction for future research.

\paragraph{Logistic Regression Knowledge Tracing Models} 

Logistic regression models of knowledge tracing such as BestLR~\cite{gervet2020deep} or LKT~\cite{pavlik2021logistic} share several similarities with VTIRT. As noted earlier, these models use the number of correct and incorrect past attempts in a learning trajectory to predict future performance, and VTIRT makes inference on ability based on both the historical performance of the learner and the features of the attempted items. While the focus of this study was to develop scalable inference for dynamic IRT models and compare the model fit against other proficiency models, it remains an interesting future research to compare VTIRT against logistic regression knowledge tracing models under both adaptive and non-adaptive learning environments.

\section{Conclusion}
\label{sec:conclusion}

We presented VTIRT, a fast and accurate inference framework for dynamic item response models. VTIRT offers orders of magnitude speedup in the inference runtime while maintaining a highly accurate inference of learner and item parameters. Moreover, every component of our inference algorithm is interpretable by virtue of its modular design. Experiments on real student data demonstrates that VTIRT achieves improvements in inferring future learner performance compared to other proficiency models.

\bibliographystyle{abbrv}
\bibliography{main}  

\begin{thebibliography}{10}

\bibitem{badrinath2021pybkt}
A.~Badrinath, F.~Wang, and Z.~Pardos.
\newblock pybkt: An accessible python library of bayesian knowledge tracing
  models.
\newblock {\em International Educational Data Mining Society}, 2021.

\bibitem{betancourt2017conceptual}
M.~Betancourt.
\newblock A conceptual introduction to hamiltonian monte carlo.
\newblock {\em arXiv preprint arXiv:1701.02434}, 2017.

\bibitem{bingham2019pyro}
E.~Bingham, J.~P. Chen, M.~Jankowiak, F.~Obermeyer, N.~Pradhan, T.~Karaletsos,
  R.~Singh, P.~Szerlip, P.~Horsfall, and N.~D. Goodman.
\newblock Pyro: Deep universal probabilistic programming.
\newblock {\em The Journal of Machine Learning Research}, 20(1):973--978, 2019.

\bibitem{brooks2011handbook}
S.~Brooks, A.~Gelman, G.~Jones, and X.-L. Meng.
\newblock {\em Handbook of markov chain monte carlo}.
\newblock CRC press, 2011.

\bibitem{choffin2019das3h}
B.~Choffin, F.~Popineau, Y.~Bourda, and J.-J. Vie.
\newblock Das3h: Modeling student learning and forgetting for optimally
  scheduling distributed practice of skills.
\newblock In {\em International Conference on Educational Data Mining (EDM
  2019)}, 2019.

\bibitem{corbett1994knowledge}
A.~T. Corbett and J.~R. Anderson.
\newblock Knowledge tracing: Modeling the acquisition of procedural knowledge.
\newblock {\em User modeling and user-adapted interaction}, 4(4):253--278,
  1994.

\bibitem{ekanadham2017t}
C.~Ekanadham and Y.~Karklin.
\newblock T-skirt: Online estimation of student proficiency in an adaptive
  learning system.
\newblock {\em arXiv preprint arXiv:1702.04282}, 2017.

\bibitem{gershman2014amortized}
S.~Gershman and N.~Goodman.
\newblock Amortized inference in probabilistic reasoning.
\newblock In {\em Proceedings of the annual meeting of the cognitive science
  society}, volume~36, 2014.

\bibitem{gervet2020deep}
T.~Gervet, K.~Koedinger, J.~Schneider, T.~Mitchell, et~al.
\newblock When is deep learning the best approach to knowledge tracing?
\newblock {\em Journal of Educational Data Mining}, 12(3):31--54, 2020.

\bibitem{hoffman2014no}
M.~D. Hoffman, A.~Gelman, et~al.
\newblock The no-u-turn sampler: adaptively setting path lengths in hamiltonian
  monte carlo.
\newblock {\em J. Mach. Learn. Res.}, 15(1):1593--1623, 2014.

\bibitem{imai2016fast}
K.~Imai, J.~Lo, and J.~Olmsted.
\newblock Fast estimation of ideal points with massive data.
\newblock {\em American Political Science Review}, 110(4):631--656, 2016.

\bibitem{johnson2016composing}
M.~J. Johnson, D.~K. Duvenaud, A.~Wiltschko, R.~P. Adams, and S.~R. Datta.
\newblock Composing graphical models with neural networks for structured
  representations and fast inference.
\newblock {\em Advances in neural information processing systems}, 29, 2016.

\bibitem{kingma2013auto}
D.~P. Kingma and M.~Welling.
\newblock Auto-encoding variational bayes.
\newblock {\em arXiv preprint arXiv:1312.6114}, 2013.

\bibitem{martin2002dynamic}
A.~D. Martin and K.~M. Quinn.
\newblock Dynamic ideal point estimation via markov chain monte carlo for the
  us supreme court, 1953--1999.
\newblock {\em Political analysis}, 10(2):134--153, 2002.

\bibitem{pavlik2021logistic}
P.~I. Pavlik, L.~G. Eglington, and L.~M. Harrell-Williams.
\newblock Logistic knowledge tracing: A constrained framework for learner
  modeling.
\newblock {\em IEEE Transactions on Learning Technologies}, 14(5):624--639,
  2021.

\bibitem{piech2015deep}
C.~Piech, J.~Bassen, J.~Huang, S.~Ganguli, M.~Sahami, L.~J. Guibas, and
  J.~Sohl-Dickstein.
\newblock Deep knowledge tracing.
\newblock {\em Advances in neural information processing systems}, 28, 2015.

\bibitem{sawyer2005cambridge}
R.~K. Sawyer.
\newblock {\em The Cambridge handbook of the learning sciences}.
\newblock Cambridge University Press, 2005.

\bibitem{studer2012incorporating}
C.~Studer.
\newblock Incorporating learning over time into the cognitive assessment
  framework.
\newblock {\em Unpublished PhD, Carnegie Mellon University, Pittsburgh, PA},
  2012.

\bibitem{van1997handbook}
W.~J. Van~der Linden and R.~Hambleton.
\newblock Handbook of item response theory.
\newblock {\em Taylor \& Francis Group. Citado na p{\'a}g}, 1(7):8, 1997.

\bibitem{van2008categorical}
P.~Van~Rijn et~al.
\newblock Categorical time series in psychological measurement.
\newblock {\em Psychometrika}, 62:215--236, 2008.

\bibitem{wang2013bayesian}
X.~Wang, J.~O. Berger, and D.~S. Burdick.
\newblock Bayesian analysis of dynamic item response models in educational
  testing.
\newblock {\em The Annals of Applied Statistics}, 7(1):126--153, 2013.

\bibitem{weng2018real}
R.~C.-H. Weng and D.~S. Coad.
\newblock Real-time bayesian parameter estimation for item response models.
\newblock {\em Bayesian Analysis}, 13(1):115--137, 2018.

\bibitem{wilson2016back}
K.~H. Wilson, Y.~Karklin, B.~Han, and C.~Ekanadham.
\newblock Back to the basics: Bayesian extensions of irt outperform neural
  networks for proficiency estimation.
\newblock {\em arXiv preprint arXiv:1604.02336}, 2016.

\bibitem{wu2020variational}
M.~Wu, R.~L. Davis, B.~W. Domingue, C.~Piech, and N.~Goodman.
\newblock Variational item response theory: Fast, accurate, and expressive.
\newblock {\em arXiv preprint arXiv:2002.00276}, 2020.

\end{thebibliography}


\appendix

\section{Proof of Theorem~\ref{thm:gaussian_potential}}
\label{sec:proof}

We will first find the parameters $\alpha_t, \beta_t, s_t$ of the resulting Linear Gaussian Model (Equation~\eqref{eq:lgm}) by solving for the following equation:
\begin{align}
    &\log q(\theta_{1:T}) \nonumber \\
    & = \left({\frac{\theta_1}{\sigma_\theta}}\right)^2 
    + \sum^T_{t=1} \left\{{\left({\frac{\theta_{t}-\theta_{t-1}}{\sigma_\theta}}\right)^2 + \left({\frac{\theta_{t}-\mu_{t}}{\sigma_t}}\right)^2}\right\}
    + C \nonumber \\
    & = \left({\frac{\theta_1 - \beta_1}{s_1}}\right)^2 +
    \sum^T_{t=2}\left({\frac{\theta_t -\alpha_{t}\theta_{t-1} - \beta_{t}}{s_t}}\right)^2 + C',
    \label{eq:equation}
\end{align}
where $C$ and $C'$ are constants with respect to $\theta_{1:T}$. Rearranging terms and comparing the coefficints of the terms involving $\theta_t\theta_{t-1}$, we obtain
\begin{equation*}
    s_t = \sigma_\theta \sqrt{\alpha_t}.
\end{equation*}
Substituting this into Equation~\eqref{eq:equation} and comparing the terms involving $\theta_t$ and $\theta_t^2$, we obtain the following recursive system of equations:
\begin{align*}
    \alpha_t &= \frac{\lambda_\theta}{\lambda_\theta + \lambda_t + (1-\alpha_{t+1})\lambda_\theta}, \\
    \beta_t &= \frac{\mu_t\lambda_t + \beta_{t+1}\lambda_\theta}{\lambda_\theta + \lambda_t + (1-\alpha_{t+1})\lambda_\theta},
\end{align*}
where $\alpha_{T+1} = 1$ and $\beta_{T+1} = 0$ are defined for notational simplicity. Note from the above equation that
\begin{equation*}
    \frac{b_t}{1-\alpha_t} 
    = \frac{\lambda_t + (1-\alpha_{t+1})\lambda_\theta}{\mu_t\lambda_t + (1-\alpha_{t+1})\lambda_\theta\left({\frac{\beta_{t+1}}{1-\alpha_{t+1}}}\right)}.
\end{equation*}
This motivates us to define $\rho_t = 1 - \alpha_t$ and $\tau_t = \frac{\beta_t}{1-\alpha_t}$, which yields the formula in Equations~\eqref{eq:rho_t} and~\eqref{eq:tau_t}:
\begin{align*}
    \rho_t &= \left({\frac{
        \lambda_t + (\rho_{t+1}\lambda_\theta)
    }{
        \lambda_\theta
        + \lambda_t + (\rho_{t+1}\lambda_\theta)
    }}\right), \;
    \tau_t = \left({\frac{
        \lambda_t\mu_t + (\rho_{t+1}\lambda_\theta)\tau_{t+1}
    }{
        \lambda_t + (\rho_{t+1}\lambda_\theta)
    }}\right).
\end{align*}

$\widetilde\mu_t$ in Equation~\eqref{eq:q_conditional} then satisfies
\begin{align*}
    \widetilde\mu_t
    &= \alpha_t\theta_{t-1} + \beta_t 
    = (1-\rho_t)\theta_{t-1} + \rho_t\tau_t \\
    &= \left({\frac{\lambda_\theta\theta_{t-1}}{\lambda_\theta + \lambda_t + (\rho_{t+1}\lambda_\theta)}}\right)
    +  \left({\frac{\lambda_t\mu_t + (\rho_{t+1}\lambda_\theta)\tau_{t+1}}{\lambda_\theta + \lambda_t + (\rho_{t+1}\lambda_\theta)}}\right) \\
    &= \left({\frac{
        \lambda_\theta\theta_{t-1}
        + \lambda_t\mu_t 
        + (\rho_{t+1}\lambda_\theta)\tau_{t+1}
     }{
        \lambda_\theta + \lambda_t 
        + (\rho_{t+1}\lambda_\theta)
     }}\right),
\end{align*}
and $\widetilde\sigma_t = s_t = \sigma_\theta\sqrt{a_t} = \sigma_\theta\sqrt{1-\rho_t}$.

\section{Experiment Details}
\label{sec:exp_details}

For all implementation of the VTIRT variants, we used a 2-layer feedforward neural network with 16 dimensional hidden layers with GELU activation for the potential function.

While TSKIRT requires the item parameters to be learned in advance using standard IRT, we used the ground-truth item parameters instead of training the item parameters with a different model - all other algorithms had to infer the item parameters from scratch.

All experiments were run on identically configured CPU machines (2 AMD EPYC 7502 32-Core Processors and 10 gigabytes of memory) until convergence for a maximum of 10 hours, \emph{with the exception of VEM}. VEM makes batch updates to the latent posterior estimates, and its item parameter updates can be significantly sped up through vectorized indexing. This speedup, however, incurs a large memory overhead. To make a conservative comparison of VTIRT's run time performance against the ideal setup for VEM, we applied this vectorization to VEM, but had to allow it to use \emph{4 times} the memory allocated to other methods, especially for the larger datasets.

\balancecolumns
\end{document}